\documentclass[10pt, twocolumn]{article}

\usepackage[utf8]{inputenc} 
\usepackage[T1]{fontenc} \usepackage{lmodern} 

\usepackage{graphicx}

\usepackage[a4paper, margin=1in]{geometry}

\usepackage[style=ieee]{biblatex}
\usepackage{multirow}

\usepackage{balance}

\usepackage{microtype}

\usepackage[font=small, labelfont=bf]{caption}

\newcommand*\samethanks[1][\value{footnote}]{\footnotemark[#1]}

\title{Towards an Automatic Diagnosis of Peripheral and Central Palsy Using Machine Learning on Facial Features}

\author{
    C.V.\ Vletter\thanks{\texttt{$\{$c.v.vletter, h.l.burger$\}$@student.hhs.nl}}
    \and 
    H.L.\ Burger\samethanks
    \and
    H.\ Alers\thanks{\texttt{ h.alers@hhs.nl}}
    \and
    N.\ Sourlos\thanks{\texttt{ n.sourlos@umcg.nl}}
    \and
    Z.\ Al-Ars\thanks{\texttt{ z.al-ars@tudelft.nl}}
}

\date{
}

\begin{document}

\maketitle

\section*{Abstract}

Central palsy is a form of facial paralysis that requires urgent medical attention and has to be differentiated from other, similar conditions such as peripheral palsy. To aid in fast and accurate diagnosis of this condition, we propose a machine learning approach to automatically classify peripheral and central facial palsy. The Palda dataset is used, which contains 103 peripheral palsy images, 40 central palsy, and 60 healthy people. Experiments are run on five machine learning algorithms. The best performing algorithms were found to be the SVM (total accuracy of 85.1\,\%) and the Gaussian naive Bayes (80.7\,\%). The lowest false negative rate on central palsy was achieved by the naive Bayes approach (80\,\% compared to 70\,\%). This condition  could prove to be the most severe, and thus its sensitivity is another good way to compare algorithms. By extrapolation, a dataset size of 334 total pictures is estimated to achieve a central palsy sensitivity of 95\,\%. All code used for these machine learning experiments is freely available online at \texttt{https://github.com/cvvletter/palsy}.

\section{Introduction}\label{sec:introduction}
Peripheral palsy and central palsy are two forms of facial paralysis. Both conditions have similar characteristics, but they can have very different acuteness. Bell’s Palsy is one form of peripheral nerve disease, from which patients often heal and which is not likely to be fatal~\cite{peitersen1982natural}. A central facial palsy, however, can be indicative of a stroke. In this case, it is vital to survival that the diagnosis is made early and that prompt treatment is provided to those patients~\cite{gordon1987dysphagia}.

\begin{figure}[t]
    \centering
    \includegraphics[width=\linewidth]{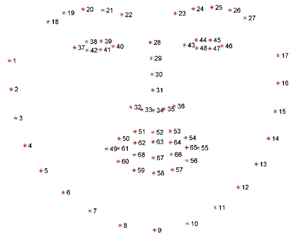}
    \caption{The location on a face of the 68 numbered facial landmarks~\cite{Saganos_annot}}
    \label{fig:introduction_landmarks_location}
\end{figure}

In earlier research~\cite{sourlos2020}, a system for an automatic palsy classification based on manually defined distances between points of the face was introduced. This system works by executing the following steps in order:
\begin{enumerate}
    \item A picture of an individual is used as input to the system, the location of the face is then found by a facial detection algorithm.
    \item A facial feature extractor finds 68 predetermined landmarks (shown in Figure~\ref{fig:introduction_landmarks_location}) on the face and records the location (x- and y-coordinates) of each one of these.
    \item A deterministic algorithm makes the diagnosis (`central palsy’, `peripheral palsy’, or `healthy’), based on predetermined distances between the landmarks and predetermined thresholds for these distances.
\end{enumerate}

The dataset used in this work, contains a total of 203 pictures found online, which were classified by an experienced neurologist and a neurologist in-training as having peripheral or central palsy, or being healthy. Due to privacy reasons, no new pictures taken in a clinical setting could be added to the dataset. Out of the 203 images, 103 pictures are of patients with peripheral palsy, 40 of patients with central palsy, and 60 of healthy people. For each picture, the coordinates of the 68 landmarks on the face, the output of step 2 of the system, are available. For patients, these are manually annotated, while for healthy people, these were placed automatically. The coordinates of the top-left and bottom-right point of a bounding box around the face region (“face box”), the output of step 1 of the system, are also available. The dataset as described above is called the “palsy dataset”.~\cite{sourlos2020}

The algorithm used in step 3 of the described system currently works by defining 49 quantitative measurements on the extracted landmarks, and choosing threshold values to distinguish between healthy people and patients, and then between a central or a peripheral facial palsy. Since all images were taken into account to define these thresholds, it was not assessed whether the algorithm works on an alternative image dataset. Therefore, it would be better to replace the current classification algorithm with a more robust one that is not based on manually defined thresholds.

The goal of this algorithm should be to classify any individual correctly. Given the data (the positions of all landmarks on a person’s face), a diagnosis must be made (healthy, central, or peripheral palsy). For these kinds of problems, a machine learning algorithm (MLA) can be considered~\cite{sajda2006machine}. Since the diagnoses for all individuals in the palsy dataset are known, a supervised learning approach is preferred~\cite{geeksforgeeks_2021}. This means the training algorithm will use both the features (data based on the landmarks) and the corresponding label (diagnosis) of any data point (image), to train the MLA.

In a similar study, researchers used pictures taken with a smartphone to detect facial palsy. For classification, the researchers used Linear Discriminant Analysis and a Support Vector Machine~\cite{kim2015smartphone}. This binary classification system, however, does not distinguish between the two possible types of palsy a patient has. 

SVMs are a type of MLA which find clusters in high-dimensional data. The data in our datasets consists of up to 136 features per data point. A total of 203 data points are available. This ratio of high complexity to small size of the palsy dataset could mean that SVMs are a good candidate for this problem.~\cite{somvanshi2016review}


More basic algorithm types of algorithms are also good to consider for this classification problem. An example could be a trained decision tree~\cite{somvanshi2016review}. Using an algorithm like ID3 to train this tree, in which metrics are used step-by-step to get closer to the final answer, could lead to an algorithm resembling the one proposed in~\cite{sourlos2020}. 

An expansion to the decision tree is the random forest. This approach uses many trees in parallel to increase higher classification certainty.~\cite{breiman2001}

\begin{figure}
    \centering
    \includegraphics[width=\linewidth]{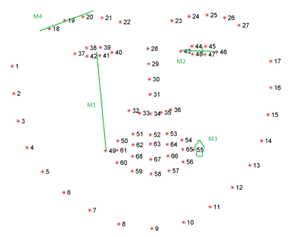}
    \caption{Example of four extracted metrics, adapted from~\cite{Saganos_annot}}
    \label{fig:method_metrics_example}
\end{figure}

\section{Materials \& methods}\label{sec:method}
This article proposes an algorithm to replace the third step of the system~\cite{sourlos2020} described in Section~\ref{sec:introduction}. To achieve this, different types of MLAs will be tested on features extracted from the palsy dataset. The accuracy of each will be recorded and later used to choose the algorithm that’s best suited for this problem. 

It should be noted that the algorithm that was previously used in the system used hand-tweaked thresholds to achieve the highest possible accuracy on the known dataset~\cite{sourlos2020}. In this article, we propose an algorithm which is tested with data not used for training. This allows for better assessment of the generalisation performance of the algorithm. Moreover, as more data become available, the algorithm can also very easily be retrained to further increase its performance.

Since the dataset is small, Leave-One-Out Cross-Validation (LOOCV) will be used to run the final algorithm accuracy tests. For a dataset of size $n$, this means that $n-1$ of the data points are used for training, and the remaining $1$ data point is used for testing. This process is repeated $n$ times, with each data point being used for testing once. The total number of correct classifications is denoted $c$. To compute the final classification accuracy $a$, the ratio of correct predictions to total predictions is used:
\begin{equation}\label{eq:method_loocv_accuracy}
    a =  \frac{c}{n} \cdot 100\,\mathrm{\%}\,.
\end{equation}

LOOCV maximises the ratio of training data to testing data, while preserving an accurate accuracy metric~\cite{kim2015smartphone}. Even though using LOOCV, it is computationally expensive to train a model, the inference time is low which makes it a great candidate for medical applications.

Another metric of interest is the classification accuracy for patients with a central palsy, since this is the most sever condition. This is called the sensitivity for this condition. For a certain condition, the sensitivity $s$ is calculated by using
\begin{equation}\label{eq:method_sensitivity}
    s = \frac{c_1}{n_1} \cdot 100\,\mathrm{\%}\,.
\end{equation}
Here, $c_1$ is the number of correct predictions for all $n_1$ number of people with a certain condition.

\subsection{Description of the datasets}\label{sub:method_datasets}
In this study, three different adaptations of the palsy dataset are used in experiments. These are described below.

\subsubsection{Landmarks dataset}\label{subsub:method_datasets_landmarks}
For supervised machine learning algorithms, it is important to have a properly annotated and labelled dataset and to perform the right pre-processing before feeding the data to those algorithms. These are the pre-processing steps performed on the palsy dataset:
\begin{enumerate}
    \item The faces are cropped from the image using the face box data, resulting from step 1 of the system described in Section~\ref{sec:introduction}.
    \item The cropped images are resized to 900x900\,pixels and the landmarks are resized accordingly.
    \item The resized images are rotated so that the mean of all points from both eyes are in the same horizontal line. The landmarks are rotated accordingly using the resulting translation matrix.
    \item Landmarks outside the face box region are shifted to the nearest point inside the face box.
    \item The resulting landmark data is then normalised, meaning all landmarks are resized so that the coordinates end up between 0 and 1.
\end{enumerate}

For all patient pictures, the manually annotated landmarks available in the palsy dataset are used. One data point was left out because many landmarks laid outside the face box, resulting in too many incorrect landmarks after step four of pre-processing. The dataset that results when applying these five steps to the palsy dataset is called the ``landmarks dataset''. It contains 202 data points, and has 136 features: two for each landmark.

\subsubsection{No-chin dataset}\label{subsub:method_datasets_nochin}
A second dataset that is experimented with, is created from the landmarks dataset described in Section~\ref{subsub:method_datasets_landmarks}. The chin points are the landmarks most often shifted by step 4 of the pre-processing steps described in Section~\ref{subsub:method_datasets_landmarks}, and they are unlikely to differ significantly between healthy people and patients. Leaving out the chin points will also decrease the processing time. Therefore, they were left out for this second dataset. This  new dataset will be called the ``no-chin dataset''. It still contains 202 data points, but the complexity is reduced to a feature size of 102: the x- and y-coordinates of the 17 chin landmarks are discarded.

\subsubsection{Metrics dataset}\label{subsub:method_datasets_metrics}
In previous research, metrics were defined based on extracted facial features which were used to give each patient a palsy score~\cite{fattah2015facial}. Two of these scores are the House-Brackmann and Sunnybrook scale. What these scales generally have in common is that they look at different parts of the face, compare them to the other side and score a patient based on the difference between the left and right size of one’s face. To make use of this, 52 metrics are defined for each of the data points, based on the landmarks dataset. The metrics are based on the House-Brackmann scale, Sunnybrook scale, and previous work~\cite{sourlos2020}. 

Some examples of facial metrics are listed here.
Ratio of distances from the centre of each eye to the edge of the corresponding side of the mouth (M1).
Ratio of the width of each eye divided by its eyelid opening (M2).
Ratio of the difference of the highest left point of the mouth and the highest right point of the mouth (M3).
Slope of the best line fitted on the 3-left most points of the left brow (M4).
These four examples are illustrated in Figure~\ref{fig:method_metrics_example}, with the labels as specified.

When the 52 metrics are calculated from the landmarks dataset, a new dataset is formed. This set will be called the ``metrics dataset''. It still contains 202 data points, but each one has 52 features.

The metrics dataset thus uses data transformation when compared to the landmarks dataset, possibly revealing distinctions between classes otherwise not easily found by a MLA.

\subsection{Description of the algorithms}\label{sub:method_algorithms}
In this work, five algorithms are tested for finding a diagnosis algorithm. These are presented below.


\subsubsection{Gaussian naive Bayes}\label{subsub:method_algorithms_naivebayes}
The Gaussian naive Bayes algorithm is based on Bayes' theorem. Naive Bayes assumes independent, equally important features and calculates a probability that a feature belongs to each class (healthy, central palsy, or peripheral palsy). It then uses the probabilities for all features to predict an outcome class. In real world situations, features often are not actually independent, yet this algorithm seems to work in practice~\cite{Chauhan_Naive}.

\subsubsection{Simple decision tree}\label{subsub:method_algorithms_decisiontree}
A simple decision tree is another supervised MLA. The decision tree decides based on multiple different features, each layer deep selecting the best feature to split the test data. Then walking the test data down a path through multiple nodes. Each of these nodes has two different edges, leading to more nodes and ending with a leaf: the outcome classification of the decision tree.

This algorithm follows a flow chart type structure that resembles human-like decision making. Because of its ability to handle both numerical and categorical features and to find all the possible outcomes~\cite{Sharma_DecisionTree}, the decision tree is a widely used MLA. 

\subsubsection{K-nearest neighbour (KNN)}\label{subsub:method_algorithms_knn}
The premises behind the KNN algorithm is rather simple. The training data points are placed in an $f$-dimensional space, where $f$ is the number of features per data point. After this, the data point to predict is placed upon the plane~\cite{Srivastava_K-nearest-neighbor}. Once the test point is placed, an $f$-dimensional sphere is drawn so that $k$ number of other data points are within the circle. Only these $k$ values will then be taken into consideration, and the test point will be assigned a label based on the majority vote amongst the $k$ points, weighed by distance.

\subsubsection{Random forest}\label{subsub:method_algorithms_randomforest}
A random forest is an extension of a simple decision tree that uses multiple uncorrelated trees, called an ensemble. The combination of these will outperform any of the individual models~\cite{Yui_RandomForest}. The idea behind the random forest it that it takes the most predicted answer and thus guarantees a higher predictability chance then a single decision tree. This is because of the effect the individual trees have on each other. While in a single decision tree a wrong answer is fatal, a random forest combats this by taking the majority vote amongst all decision tree predictions. 


\subsubsection{Support vector machine}\label{subsub:method_algorithms_svm}
The idea of a support vector machine is that a hyperplane which separates the data into classes is created~\cite{Pupale_SVM}. The training algorithm looks for the margin between both classes and places a hyperplane so that the margin is maximised

When a perfectly separating hyperplane cannot be created, the data can be transformed into a higher dimension using a kernel function. Possibly, a hyperplane can now be fitted into the plot. There are several different kernel functions. 



\section{Experimental results}\label{sec:results}
To test which algorithm performs best in the task of diagnosing any person as being healthy, as having a peripheral or as having a central palsy, experiments are run with the five algorithms discussed in Section~\ref{sub:method_algorithms}. From the Scikit-learn machine learning library~\cite{pedregosa2011scikit}, the algorithms SVM, random forest, Gaussian naive Bayes, KNN, and decision tree are used. 
The dataset, the Python files to create the three  and Jupyter notebooks used to run all experiments are available on Github\footnote{\texttt{https://github.com/cvvletter/palsy}}. The experiments are run on the three datasets: the landmarks dataset, the no-chin dataset, and the metrics dataset (explained in Section~\ref{sub:method_datasets}).

For each algorithm, excluding the neural network, the LOOCV accuracies $a$ from (\ref{eq:method_loocv_accuracy}) are given for all three datasets. Some of the experiment runs are accompanied by a confusion matrix. These matrices (which all look like Table~\ref{tab:results_confusionmatrix_naivebayes_nochin}) show the type of error the algorithm makes most. For each actual diagnosis, the percentages of predicted diagnoses are stated. In these tables, abbreviations are used for the three classes. P means peripheral palsy, C means central palsy, and H means healthy. If an algorithm performs very good, the top-left to bottom-right diagonal will contain only values close to or at 100\,\%, and all other cells will contain values close to or at 0\,\%.



\subsection{Gaussian naive Bayes performance}\label{sub:results_naivebayes_performance}
For the Gaussian naive Bayes algorithm, the results are as follows: When using the landmarks dataset in training the model, an accuracy of $a=63.9\,\mathrm{\%}$ is achieved. The accuracy improves to $a=64.4\,\mathrm{\%}$ when the no-chin dataset is used. For the no-chin dataset, a confusion matrix is shown in Table~\ref{tab:results_confusionmatrix_naivebayes_nochin}.

The Gaussian naive Bayes ML algorithm was then fitted on the metrics dataset. A testing accuracy of $a=80.7\,\mathrm{\%}$ is now reached. The resulting confusion matrix is given in Table~\ref{tab:results_confusionmatrix_naivebayes_metrics}. The Gaussian naive Bayes model has no parameters that can be tuned.

\begin{table}[bp]
    \centering
    \caption{Confusion matrix for the Gaussian naive Bayes algorithm on the no-chin dataset}
    \label{tab:results_confusionmatrix_naivebayes_nochin}
    \begin{tabular}{|c|c|ccc|}
    \cline{3-5}
    \multicolumn{2}{c|}{\multirow{2}{*}{Diagnosis}} & \multicolumn{3}{|c|}{Actual} \\
    \cline{3-5}
    \multicolumn{2}{c|}{} & P & C & H \\
    \hline
     & P & 69.6\,\% & 40.0\,\% & 18.3\,\% \\
     & C & 28.4\,\% & 40.0\,\% & 10.0\,\% \\
    \multirow{-3}{*}{Predicted} & H & 2.0\,\% & 20.0\,\% & 71.7\,\% \\
    \hline
    \end{tabular}
\end{table}

\begin{table}[bp]
    \centering
    \caption{Confusion matrix for the Gaussian naive Bayes algorithm on the metrics dataset} 
    \label{tab:results_confusionmatrix_naivebayes_metrics}
    \begin{tabular}{|c|c|ccc|}
    \cline{3-5}
    \multicolumn{2}{c|}{\multirow{2}{*}{Diagnosis}} & \multicolumn{3}{|c|}{Actual} \\
    \cline{3-5}
    \multicolumn{2}{c|}{} & P & C & H \\
    \hline
     & P & 73.5\,\% & 20.0\,\% & 1.7\,\% \\
     & C & 26.5\,\% & 80.0\,\% & 5.0\,\% \\
    \multirow{-3}{*}{Predicted} & H & 0\,\% & 0\,\% & 93.3\,\% \\
    \hline
    \end{tabular}
\end{table}

\subsection{Decision tree performance}\label{sub:results_decisiontree_performance}
A decision tree model is also built. Using a max depth of 10 for the tree, accuracies of $a=67.8\,\mathrm{\%}$ and $a=69.8\,\mathrm{\%}$ are achieved, if using the landmarks and no-chin datasets, respectively. For the no-chin dataset, the confusion matrix given in Table~\ref{tab:results_confusionmatrix_decisiontree_nochin} results.

The simple decision tree reached an accuracy of $a=73.3\,\mathrm{\%}$ on the metrics dataset when a maximum tree depth of 20 is used. The resulting confusion matrix is given in Table~\ref{tab:results_confusionmatrix_decisiontree_metrics}. Besides the tree-depth, no further parameter tuning was performed.

\begin{table}[bp]
    \centering
    \caption{Confusion matrix for the decision tree algorithm on the no-chin dataset}
    \label{tab:results_confusionmatrix_decisiontree_nochin}
    \begin{tabular}{|c|c|ccc|}
    \cline{3-5}
    \multicolumn{2}{c|}{\multirow{2}{*}{Diagnosis}} & \multicolumn{3}{|c|}{Actual} \\
    \cline{3-5}
    \multicolumn{2}{c|}{} & P & C & H \\
    \hline
     & P & 81.4\,\% & 35.0\,\% & 13.3\,\% \\
     & C & 11.8\,\% & 47.5\,\% & 21.7\,\% \\
    \multirow{-3}{*}{Predicted} & H & 6.7\,\% & 17.5\,\% & 65.0\,\% \\
    \hline
    \end{tabular}
\end{table}

\begin{table}[bp]
    \centering
    \caption{Confusion matrix for the decision tree algorithm on the metrics dataset}
    \label{tab:results_confusionmatrix_decisiontree_metrics}
    \begin{tabular}{|c|c|ccc|}
    \cline{3-5}
    \multicolumn{2}{c|}{\multirow{2}{*}{Diagnosis}} & \multicolumn{3}{|c|}{Actual} \\
    \cline{3-5}
    \multicolumn{2}{c|}{} & P & C & H \\
    \hline
     & P & 75.5\,\% & 45.0\,\% & 5.0\,\% \\
     & C & 19.6\,\% & 42.5\,\% & 5.0\,\% \\
    \multirow{-3}{*}{Predicted} & H & 6.7\,\% & 12.5\,\% & 90.0\,\% \\
    \hline
    \end{tabular}
\end{table}

\subsection{K-nearest neighbour performance}\label{sub:results_knn_performance}
For the K-nearest neighbour approach, the model was specified to use the nearest 5 neighbours to the data point to predict the output class. These neighbours were weighed by distance, meaning closer neighbours count more in the final prediction. When using this approach, the model achieves an accuracy of $a=69.8\,\mathrm{\%}$ on the landmark’s dataset. When the no-chin dataset is used, the accuracy is improved to $a=75.2\,\mathrm{\%}$. The confusion matrix for the no-chin dataset is shown in Table~\ref{tab:results_confusionmatrix_knn_nochin}.

When the nearest 7 neighbours are considered and again weighed by distance, a classification accuracy of $a=59.4\,\mathrm{\%}$ is achieved on the metrics dataset. The confusion matrix that results is shown in Table~\ref{tab:results_confusionmatrix_knn_metrics}.

\begin{table}[bp]
    \centering
    \caption{Confusion matrix for the KNN algorithm on the no-chin dataset}
    \label{tab:results_confusionmatrix_knn_nochin}
    \begin{tabular}{|c|c|ccc|}
    \cline{3-5}
    \multicolumn{2}{c|}{\multirow{2}{*}{Diagnosis}} & \multicolumn{3}{|c|}{Actual} \\
    \cline{3-5}
    \multicolumn{2}{c|}{} & P & C & H \\
    \hline
     & P & 83.3\,\% & 42.5\,\% & 3.3\,\% \\
     & C & 8.8\,\% & 27.5\,\% & 3.3\,\% \\
    \multirow{-3}{*}{Predicted} & H & 7.9\,\% & 30.0\,\% & 93.4\,\% \\
    \hline
    \end{tabular}
\end{table}

\begin{table}[bp]
    \centering
    \caption{Confusion matrix for the KNN algorithm on the metrics dataset}
    \label{tab:results_confusionmatrix_knn_metrics}
    \begin{tabular}{|c|c|ccc|}
    \cline{3-5}
    \multicolumn{2}{c|}{\multirow{2}{*}{Diagnosis}} & \multicolumn{3}{|c|}{Actual} \\
    \cline{3-5}
    \multicolumn{2}{c|}{} & P & C & H \\
    \hline
     & P & 68.6\,\% & 57.5\,\% & 18.3\,\% \\
     & C & 13.7\,\% & 20.0\,\% & 11.7\,\% \\
    \multirow{-3}{*}{Predicted} & H & 17.6\,\% & 22.5\,\% & 70.0\,\% \\
    \hline
    \end{tabular}
\end{table}

\subsection{Random forest performance}\label{sub:results_randomforest_performance}
The random forest model has a parameter that defines the number of parallel trees. This parameter was tuned before testing the accuracy of the model.

\subsubsection{Random forest tuning}\label{subsub:results_randomforest_performance_tuning}
A random forest algorithm was also trained. The number of estimators was initially set to 136 random estimators, and used compare the landmarks and no-chin datasets. Using the no-chin dataset led to the same accuracy as when the landmarks set was used.

\begin{figure}[tp]
    \centering
    \includegraphics[width=\linewidth]{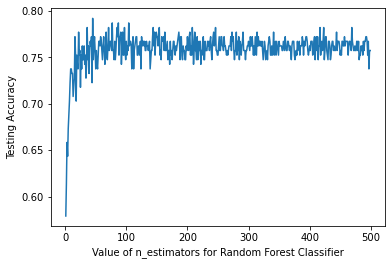}
    \caption{Random forest accuracy for different number of random estimators using the landmarks dataset}
    \label{fig:results_randomforest_tuning_landmarks}
\end{figure}

An experiment is then performed to find the optimal number of random estimators, using the landmarks dataset. A graph showing the LOOCV accuracy $a$ and number of random estimators is given in Figure~\ref{fig:results_randomforest_tuning_landmarks}. From around 100 random estimators, there seems to be no more improvement in accuracy. The `noise' – due to the randomness in the algorithm – seems to stabilise from around 150 random estimators. The training time also increases rapidly when introducing more random estimators.

\begin{figure}[tp]
    \centering
    \includegraphics[width=\linewidth]{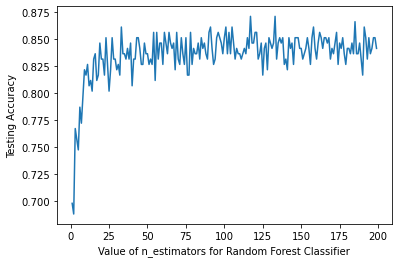}
    \caption{Random forest accuracy for different number of random estimators using the metrics dataset}
    \label{fig:results_randomforest_tuning_metrics}
\end{figure}

On the metrics dataset, the random forest algorithm was also tested with 1 until 200 random estimators. As seen in Figure~\ref{fig:results_randomforest_tuning_metrics}, no meaningful improvement in accuracy is achieved when more than around 50 random estimators are used to classify the input.

\subsubsection{Random forest testing}\label{subsub:results_randomforest_performance_testing}
A final accuracy of $a=77.2\,\mathrm{\%}$ is obtained on the landmarks dataset when 200 random estimators are used. Table~\ref{tab:results_confusionmatrix_randomforest_landmarks} shows the resulting confusion matrix for this experiment.

At last, an experiment was performed on the metrics dataset using 100 random estimators. This resulted in an accuracy of $a=85.1\,\mathrm{\%}$, with the confusion matrix given in Table~\ref{tab:results_confusionmatrix_randomforest_metrics}.

\begin{table}[bp]
    \centering
    \caption{Confusion matrix for the random forest algorithm on the landmarks dataset}
    \label{tab:results_confusionmatrix_randomforest_landmarks}
    \begin{tabular}{|c|c|ccc|}
    \cline{3-5}
    \multicolumn{2}{c|}{\multirow{2}{*}{Diagnosis}} & \multicolumn{3}{|c|}{Actual} \\
    \cline{3-5}
    \multicolumn{2}{c|}{} & P & C & H \\
    \hline
     & P & 91.2\,\% & 42.5\,\% & 10.0\,\% \\
     & C & 5.9\,\% & 22.5\,\% & 0\,\% \\
    \multirow{-3}{*}{Predicted} & H & 2.9\,\% & 7.5\,\% & 90.0\,\% \\
    \hline
    \end{tabular}
\end{table}

\begin{table}[bp]
    \centering
    \caption{Confusion matrix for the random forest algorithm on the metrics dataset}
    \label{tab:results_confusionmatrix_randomforest_metrics}
    \begin{tabular}{|c|c|ccc|}
    \cline{3-5}
    \multicolumn{2}{c|}{\multirow{2}{*}{Diagnosis}} & \multicolumn{3}{|c|}{Actual} \\
    \cline{3-5}
    \multicolumn{2}{c|}{} & P & C & H \\
    \hline
     & P & 95.1\,\% & 55.0\,\% & 0\,\% \\
     & C & 4.9\,\% & 37.5\,\% & 0\,\% \\
    \multirow{-3}{*}{Predicted} & H & 0\,\% & 7.5\,\% & 100\,\% \\
    \hline
    \end{tabular}
\end{table}

\subsection{Support vector machine performance}\label{sub:results_svm_performance}
The support vector machine was also first tuned, before testing the final accuracy.

\subsubsection{Support vector machine tuning}\label{subsub:results_svm_performance_tuning}
For training the SVM, the input classes are balanced, meaning the three classes are given a weight inversely proportional to the number of entries in that class to combat underrepresentation. The standard available kernels are used in the experiments. Both the radial basis function kernel and the linear kernel result in lower accuracy for the model than when a polynomial kernel of degree~5 is used. The polynomial kernel function is thus further examined.

\begin{figure}[tp]
    \centering
    \includegraphics[width=\linewidth]{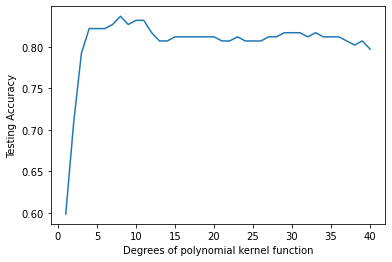}
    \caption{SVM accuracy for different degrees of polynomial kernel function using the landmarks dataset}
    \label{fig:results_svm_performance_tuning_landmarks}
\end{figure}

\begin{figure}[tp]
    \centering
    \includegraphics[width=\linewidth]{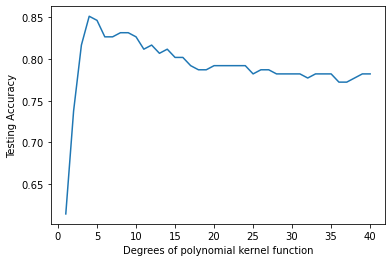}
    \caption{SVM accuracy for different degrees of polynomial kernel function using the no-chin dataset}
    \label{fig:results_svm_performance_tuning_nochin}
\end{figure}

To test the optimal degree of kernel polynomial function, values of~1 until~40 are tested using both the landmarks (Figure~\ref{fig:results_svm_performance_tuning_landmarks}) and the no-chin (Figure~\ref{fig:results_svm_performance_tuning_nochin}) datasets.

\subsubsection{Support vector machine testing}\label{subsub:results_svm_performance_testing}
Figures~\ref{fig:results_svm_performance_tuning_landmarks} and~\ref{fig:results_svm_performance_tuning_nochin} show that the highest accuracy is achieved using the no-chin dataset and a polynomial kernel function of degree 4. The final accuracy for this SVM is $a=85.1\,\mathrm{\%}$. The confusion matrix for this dataset is given in Table~\ref{tab:results_confusionmatrix_svm_nochin}.

Finally, the support vector machine was tested on the metrics dataset. The polynomial kernel function and balanced class weights still performed best, but a final accuracy of only $a=61.9\,\mathrm{\%}$ was achieved, using a polynomial kernel function of degree~15. The resulting confusion matrix is given in Table~\ref{tab:results_confusionmatrix_svm_metrics}.

\begin{table}[bp]
    \centering
    \caption{Confusion matrix for the SVM algorithm on the no-chin dataset}
    \label{tab:results_confusionmatrix_svm_nochin}
    \begin{tabular}{|c|c|ccc|}
    \cline{3-5}
    \multicolumn{2}{c|}{\multirow{2}{*}{Diagnosis}} & \multicolumn{3}{|c|}{Actual} \\
    \cline{3-5}
    \multicolumn{2}{c|}{} & P & C & H \\
    \hline
     & P & 82.4\,\% & 30.0\,\% & 0\,\% \\
     & C & 17.7\,\% & 70.0\,\% & 0\,\% \\
    \multirow{-3}{*}{Predicted} & H & 0\,\% & 0\,\% & 100\,\% \\
    \hline
    \end{tabular}
\end{table}

\begin{table}[bp]
    \centering
    \caption{Confusion matrix for the SVM algorithm on the metrics dataset}
    \label{tab:results_confusionmatrix_svm_metrics}
    \begin{tabular}{|c|c|ccc|}
    \cline{3-5}
    \multicolumn{2}{c|}{\multirow{2}{*}{Diagnosis}} & \multicolumn{3}{|c|}{Actual} \\
    \cline{3-5}
    \multicolumn{2}{c|}{} & P & C & H \\
    \hline
     & P & 49.0\,\% & 15.0\,\% & 0\,\% \\
     & C & 37.3\,\% & 57.5\,\% & 13.3\,\% \\
    \multirow{-3}{*}{Predicted} & H & 13.7\,\% & 27.5\,\% & 86.7\,\% \\
    \hline
    \end{tabular}
\end{table}

\subsection{Varying the dataset size}\label{sub:results_datasetsize_experiments}
\begin{figure}[tp]
    \centering
    \includegraphics[width=\linewidth]{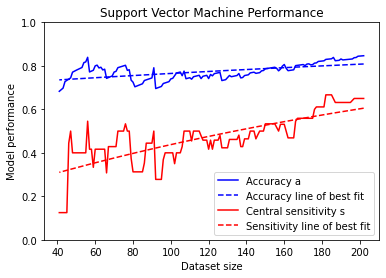}
    \caption{SVM accuracy $a$ and sensitivity $s$ for central palsy for different dataset sizes on the landmarks dataset}
    \label{fig:results_datasetsize_svm}
\end{figure}

\begin{figure}[tp]
    \centering
    \includegraphics[width=\linewidth]{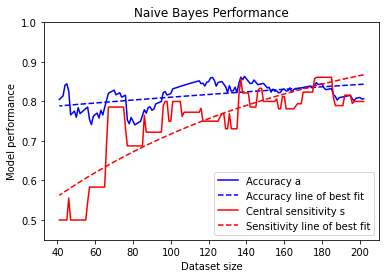}
    \caption{Gaussian naive Bayes accuracy $a$ and sensitivity $s$ for central palsy for different dataset sizes on the landmarks dataset}
    \label{fig:results_datasetsize_naivebayes}
\end{figure}

Finally, the accuracy $a$ and central palsy sensitivity $s$ are investigated when the dataset varies in size. Data points are removed one at a time from the dataset. For every 10 data points removed, 5 are peripheral palsy patients, 3 are central palsy patients, and 2 are healthy people. This roughly keeps the same division between data points in the dataset. The accuracy and sensitivity are then calculated until the dataset has only 40 data points left --$20\,\mathrm{\%}$ of the original dataset. The resulting plots for the SVM on the landmarks dataset and the Gaussian naive Bayes on the metrics dataset are shown in Figures~\ref{fig:results_datasetsize_svm} and~\ref{fig:results_datasetsize_naivebayes}, respectively.

The behaviour at larger datasets is then estimated by fitting the following curve to all lines:
\[ y = f(x) = 1 - a \cdot e^{b \cdot (x - c)} \,. \]
Here, $a$, $b$, and $c$ are unknown coefficients, $y$ and $x$ the dependent and independent variable, and $f(x)$ is a function that approaches 1 at $x \to \infty$. This is reasonable, since we would expect a machine learning algorithm to perform perfectly at infinite training dataset size.

The resulting curve fit lines are also shown in Figures~\ref{fig:results_datasetsize_svm} and~\ref{fig:results_datasetsize_naivebayes}. We highlight the line fitted to the naive Bayes sensitivity, which is as follows:
\[ y = 1 - 2.64 \cdot e^{-0.00741 \cdot (x + 201)} \,. \]
This line would intersect a performance of $95\,\mathrm{\%}$ at a dataset size of 334 pictures.

\section{Discussion}\label{sec:discussion}
The four best performing algorithms were the Gaussian naive Bayes on the metrics dataset (80.7\,\%), the random forest on the metrics dataset (85.1\,\%), and the support vector machine on the no-chin dataset (85.1\,\%). The confusion matrices of these algorithms are given in Tables~\ref{tab:results_confusionmatrix_naivebayes_metrics}, \ref{tab:results_confusionmatrix_randomforest_metrics}, and~\ref{tab:results_confusionmatrix_svm_nochin}, respectively. 

Comparing these confusion matrices, we find that while the random forest achieved the highest accuracy, it correctly classifies individuals with a central palsy only in 37.5\,\% of cases, which is not a good performance. It does perform well on healthy individuals (100\,\%) and people with a peripheral facial palsy (95.1\,\%). Since a central palsy could be a stroke, the poor performance on this condition is not ideal. This result could be caused by the unbalance of data points between the three classes. 

The SVM has the highest performance on healthy people, since it correctly classifies 100\,\% of them as healthy, and also classifies 0\,\% of patients as being healthy. The highest central palsy accuracy is reached by the naive Bayes algorithm. This algorithm classified 80\,\% of central palsy patients correctly, and thus `misses' the least cases of this disease. It could thus be argued that this algorithm performs the best, even though the random forest and SVM both reached higher overall accuracies $a$.

Another noteworthy result is that five out of six algorithms --only the random forest did not-- performed better on the no-chin dataset than on the landmarks dataset, even though the no-chin dataset is just the landmark dataset with the features corresponding to 17 landmarks left out. We assume this to be caused by the fact that the chin position differs very little between healthy people and patients. This could result in a distorted view of the differences between the three classes, which leads to poorer clustering abilities of the algorithms, and thus a lower testing accuracy.

Lastly, Figure~\ref{fig:results_datasetsize_svm} and~\ref{fig:results_datasetsize_naivebayes} show steady increase in both the general accuracy and sensitivity of central palsy. The Gaussian naive Bayes however, shows a steeper slope on sensitivity. Adding more images to the dataset could give a superior central prediction earlier on compared to the SVM approach. By curve fitting, an expected dataset size of 334 images is needed to achieve a central palsy sensitivity of 95\,\% with the Gaussian naive Bayes approach.


\section{Conclusions}\label{sub:conclusionsrecommendations_conclusions}
The goal of this study was to implement an algorithm to classify any individual as healthy or as having a peripheral or central palsy, by using the relative location of 68 facial features (landmarks) on their face. In machine learning, this is described as a three-way classification problem. The dataset used in our study contains 202~individuals.

Experiments were then run with five different algorithm types, explained in Section~\ref{sub:method_algorithms}, on three different representations of the palsy dataset, as described in Section~\ref{sub:method_datasets}. Some algorithms were first optimised by varying some of their model parameters and running accuracy tests for each.

The highest accuracies that were reached are 85.1\,\% using a SVM with a polynomial kernel function of degree~4 on the landmarks dataset and 85.1\,\% using a random forest with 100~random estimators on the metrics dataset. Both made no errors when an individual is healthy. The random forest on the metrics data seems to be overfitted to peripheral patients (see Table~\ref{tab:results_confusionmatrix_randomforest_metrics}), while the SVM on the landmarks dataset performs similarly for both palsies (see Table~\ref{tab:results_confusionmatrix_svm_nochin}). The highest correct classification rate for patients with a central palsy –-the most severe condition-– was reached by the Gaussian naive Bayes on the metrics dataset (80\,\%), while achieving a lower accuracy than the SVM and random forest of 80.7\,\%.

By curve fitting, it was shown that for both the SVM and naive Bayes algorithms, accuracy $a$ and central palsy sensitivity $s$ tend to improve when the dataset grows in size. It was estimated that by tripling the dataset in size, central palsy sensitivity would achieve 95\,\% with the naive Bayes approach.

\section{Recommendations}\label{sub:conclusionsrecommendations_recommendations}
Adding more images to the dataset would be a first step to improving the current results. This could allow the training of a deep neural network (DNN), which could achieve much better results.

Further tuning of the different MLAs to improve the accuracy is also advised. Some model parameters of MLAs presented in this study have not yet been tuned. It is expected that tweaking some of these could still lead to better results. 

Another point that needs more research, is the weight of the two palsy conditions. Since central palsy is a much more severe disease than peripheral palsy, it could be preferred that an automatic algorithm favours central palsy diagnosis. For central palsy, a false positive is less dangerous than a false negative, since the first would just alert a doctor, while the second could falsely ease one. In short, sensitivity and specificity of central palsy diagnosis could be another metric – besides accuracy – to compare different MLAs.

The significance of each of the 68 landmarks could also use more research. Since leaving out the chin landmarks resulted in higher accuracy for five out of six landmarks, there is reason to believe that some other combination of landmarks could also have this effect.

Furthermore, the line fitting on the performance curves (Figures~\ref{fig:results_datasetsize_svm} and~\ref{fig:results_datasetsize_naivebayes}) could be further investigated. The negative exponent curve we fitted to all lines might not be perfect. It could for example be investigated if a different base number than Euler's number achieves better results.

Lastly, a platform should be made to easily implement the diagnosis system in a hospital setting, after it has surpassed the preferred accuracy. Fitting a linear, polynomial, or a curve fitted regression model on the dataset-size accuracy discussed in Section~\ref{sub:results_datasetsize_experiments} should result in a prediction model to predict the preferred accuracy.  For this, steps~1 and~2 as described in Section~\ref{sec:introduction} should be combined with a diagnosis algorithm proposed in this article to form a complete automatic palsy diagnosis system.


\balance

\printbibliography

@unpublished{sourlos2020,
  title={Palda: Public Dataset and Algorithms for Facial Imaging and Diagnosis of Neurological Disorders},
  author={Sourlos, N and Amerika, W E and Jaber, Z and Jacobs, B C and de Bruijn, S F T M and Al-Ars, Z},
  note={Note: not yet published},
  year={2022},
}

@article{breiman2001,
  title={Random forests},
  author={Breiman, Leo},
  journal={Machine learning},
  volume={45},
  number={1},
  pages={5--32},
  year={2001},
  publisher={Springer},
}

@article{peitersen1982natural,
  title={The natural history of Bell's palsy},
  author={Peitersen, Erik},
  journal={The American Journal of Otology},
  volume={4},
  number={2},
  pages={107--111},
  year={1982}
}

@article{gordon1987dysphagia,
  title={Dysphagia in acute stroke},
  author={Gordon, Caroline and Hewer, R Langton and Wade, Derick T},
  journal={Br Med J (Clin Res Ed)},
  volume={295},
  number={6595},
  pages={411--414},
  year={1987},
  publisher={British Medical Journal Publishing Group}
}

@misc{Saganos_annot,
  author={Saganos, C},
  title={The 68-points mark-up used for annotation},
  year={2021},
  url={https://ibug.doc.ic.ac.uk/resources/facial-point-annotations/}
}

@article{sajda2006machine,
  title={Machine learning for detection and diagnosis of disease},
  author={Sajda, Paul},
  journal={Annu. Rev. Biomed. Eng.},
  volume={8},
  pages={537--565},
  year={2006},
  publisher={Annual Reviews}
}

@misc{geeksforgeeks_2021, 
  title={Supervised and unsupervised learning}, 
  url={https://www.geeksforgeeks.org/supervised-unsupervised-learning/}, 
  journal={GeeksforGeeks}, 
  year={2021}, 
  month={10}
}

@article{kim2015smartphone,
  title={A smartphone-based automatic diagnosis system for facial nerve palsy},
  author={Kim, Hyun Seok and Kim, So Young and Kim, Young Ho and Park, Kwang Suk},
  journal={Sensors},
  volume={15},
  number={10},
  pages={26756--26768},
  year={2015},
  publisher={Multidisciplinary Digital Publishing Institute}
}

@inproceedings{somvanshi2016review,
  title={A review of machine learning techniques using decision tree and support vector machine},
  author={Somvanshi, Madan and Chavan, Pranjali and Tambade, Shital and Shinde, SV},
  booktitle={2016 international conference on computing communication control and automation (ICCUBEA)},
  pages={1--7},
  year={2016},
  organization={IEEE}
}

@article{fattah2015facial,
  title={Facial nerve grading instruments: systematic review of the literature and suggestion for uniformity},
  author={Fattah, Adel Y and Gurusinghe, Anthony DR and Gavilan, Javier and Hadlock, Tessa A and Marcus, Jeff R and Marres, Henri and Nduka, Charles C and Slattery, William H and Snyder-Warwick, Alison K and others},
  journal={Plastic and reconstructive surgery},
  volume={135},
  number={2},
  pages={569--579},
  year={2015},
  publisher={LWW}
}

@misc{Sharma_DecisionTree,
  author={Sharma},
  title={Decision Tree Classification | Guide to Decision Tree Classification Analytics Vidhya},
  url={https://www.analyticsvidhya.com/blog/2021/04/beginners-guide-to-decision-tree-classification-using-python/},
  year={2021}
}

@misc{Srivastava_K-nearest-neighbor,
  author={Srivastava, T},
  title={K Nearest Neighbor | KNN Algorithm | KNN in Python \& R. Analytics Vidhya.},
  url={https://www.analyticsvidhya.com/blog/2018/03/introduction-k-neighbours-algorithm-clustering/},
  year={2018}
}

@misc{Yui_RandomForest,
  author={Yui, T},
  title={Understanding Random Forest - Towards Data Science},
  url={https://towardsdatascience.com/understanding-random-forest-58381e0602d2},
  year={2021}
}

@misc{Pupale_SVM,
  author={Pupale, R},
  title={Support Vector Machines (SVM) - An Overview - Towards Data Science},
  url={https://towardsdatascience.com/https-medium-com-pupalerushikesh-svm-f4b42},
  year={2019}
}

@article{pedregosa2011scikit,
  title={Scikit-learn: Machine learning in Python},
  author={Pedregosa, Fabian and Varoquaux, Ga{\"e}l and Gramfort, Alexandre and Michel, Vincent and Thirion, Bertrand and Grisel, Olivier and Blondel, Mathieu and Prettenhofer, Peter and Weiss, Ron and Dubourg, Vincent and others},
  journal={the Journal of machine Learning research},
  volume={12},
  pages={2825--2830},
  year={2011},
  publisher={JMLR. org}
}

@misc{Chauhan_Naive,
  title={Naive Bayes Algorithm: Everything you need to know},
  author={Chauhan, H. S.},
  url={https://www.kdnuggets.com/2020/06/naive-bayes-algorithm-everything.html},
  year={2020}
}

\end{document}